\documentclass[journal]{IEEEtran}
\usepackage{graphicx}
\usepackage{amsmath}
\usepackage{booktabs}
\usepackage{bm}
\usepackage{amssymb}
\usepackage{longtable}
\usepackage{multirow}
\usepackage{subfigure}
 \usepackage{array}

\usepackage{algorithm}
\usepackage{algorithmic}
\usepackage{textcomp}
\usepackage{setspace}
\DeclareMathOperator*{\argmin}{arg min}

\ifCLASSINFOpdf
\else
\fi
\begin{document}
%
\title{AdapNet: Adaptability Decomposing Encoder-Decoder Network for Weakly Supervised Action Recognition and Localization}
%
%
%

\author{Xiao-Yu Zhang,~\IEEEmembership{Senior Member,~IEEE,}
        Changsheng Li,~\IEEEmembership{Member,~IEEE,} Haichao Shi,
        Xiaobin Zhu, \\ Peng Li, Jing Dong,~\IEEEmembership{Senior Member,~IEEE}
\thanks{X.-Y Zhang and H. Shi are with Institute of Information Engineering, Chinese Academy of Sciences, Beijing, China, 100093. H. Shi is also with School of Cyber Security, University of Chinese Academy of Sciences, Beijing, China. (e-mail: zhangxiaoyu@iie.ac.cn; shihaichao@iie.ac.cn)}
\thanks{C. Li is with University of Electronic Science and Technology of China, Chengdu, China. (e-mail: lichangsheng@uestc.edu.cn)}
\thanks{X. Zhu is with University of Science and Technology Beijing, Beijing, China. (e-mail: brucezhucas@gmail.com)}
\thanks{P. Li is with College of Information and Control Engineering, China University of Petroleum (East China), 66 West Changjiang Road, Qingdao, China. (e-mail: lipeng@upc.edu.cn)}
\thanks{J. Dong is with Center for Research on Intelligent Perception and Computing, National Laboratory of Pattern Recognition, Institute of Automation, Chinese Academy of Sciences, Beijing, China. (e-mail: jdong@nlpr.ia.ac.cn)}
}

\maketitle

\begin{abstract}
The point process is a solid framework to model sequential data, such as videos, by exploring the underlying relevance. As a challenging problem for high-level video understanding, weakly supervised action recognition and localization in untrimmed videos has attracted intensive research attention. Knowledge transfer by leveraging the publicly available trimmed videos as external guidance is a promising attempt to make up for the coarse-grained video-level annotation and improve the generalization performance. However, unconstrained knowledge transfer may bring about irrelevant noise and jeopardize the learning model. This paper proposes a novel adaptability decomposing encoder-decoder network to transfer reliable knowledge between trimmed and untrimmed videos for action recognition and localization via bidirectional point process modeling, given only video-level annotations. By decomposing the original features into domain-adaptable and domain-specific ones based on their adaptability, trimmed-untrimmed knowledge transfer can be safely confined within a more coherent subspace. An encoder-decoder based structure is carefully designed and jointly optimized to facilitate effective action classification and temporal localization. Extensive experiments are conducted on two benchmark datasets (i.e., THUMOS14 and ActivityNet1.3), and experimental results clearly corroborate the efficacy of our method.
\end{abstract}

\begin{IEEEkeywords}
Action Recognition, Temporal Action Localization, Encoder-decoder, Knowledge Transfer, Point Process.
\end{IEEEkeywords}

%
\IEEEpeerreviewmaketitle

\section{Introduction}
Action recognition and temporal localization in untrimmed videos is a challenging problem with widespread attention in machine learning and computer vision communities~\cite{ADHM,MGM,CDA,VSD}. Compared with trimmed videos which are precisely segmented and well aligned with specific actions, untrimmed videos are typically cluttered with both actions and irrelevant backgrounds, and thus far more complicated to handle. Recent progress in action recognition and localization of untrimmed videos mainly concentrates on the fully supervised setting, which requires frame-level annotations or equivalently temporal intervals of actions. Considering the relatively long duration and complex content of untrimmed videos, manually labeling such fine-grained information is prohibitively labor-intensive and time-consuming. Therefore, the fully supervised methods may not be applicable to large scale untrimmed video sets. This motivates us to develop efficient learning methods with minimal demand for labeling effort. To this end, we focus on the weakly supervised scenario, which only requires the comparatively coarse video-level labels in the training stage, and still aims to identify actions and the corresponding temporal intervals in untrimmed videos during testing. 

\begin{figure}
\centering
\includegraphics[width=1\linewidth]{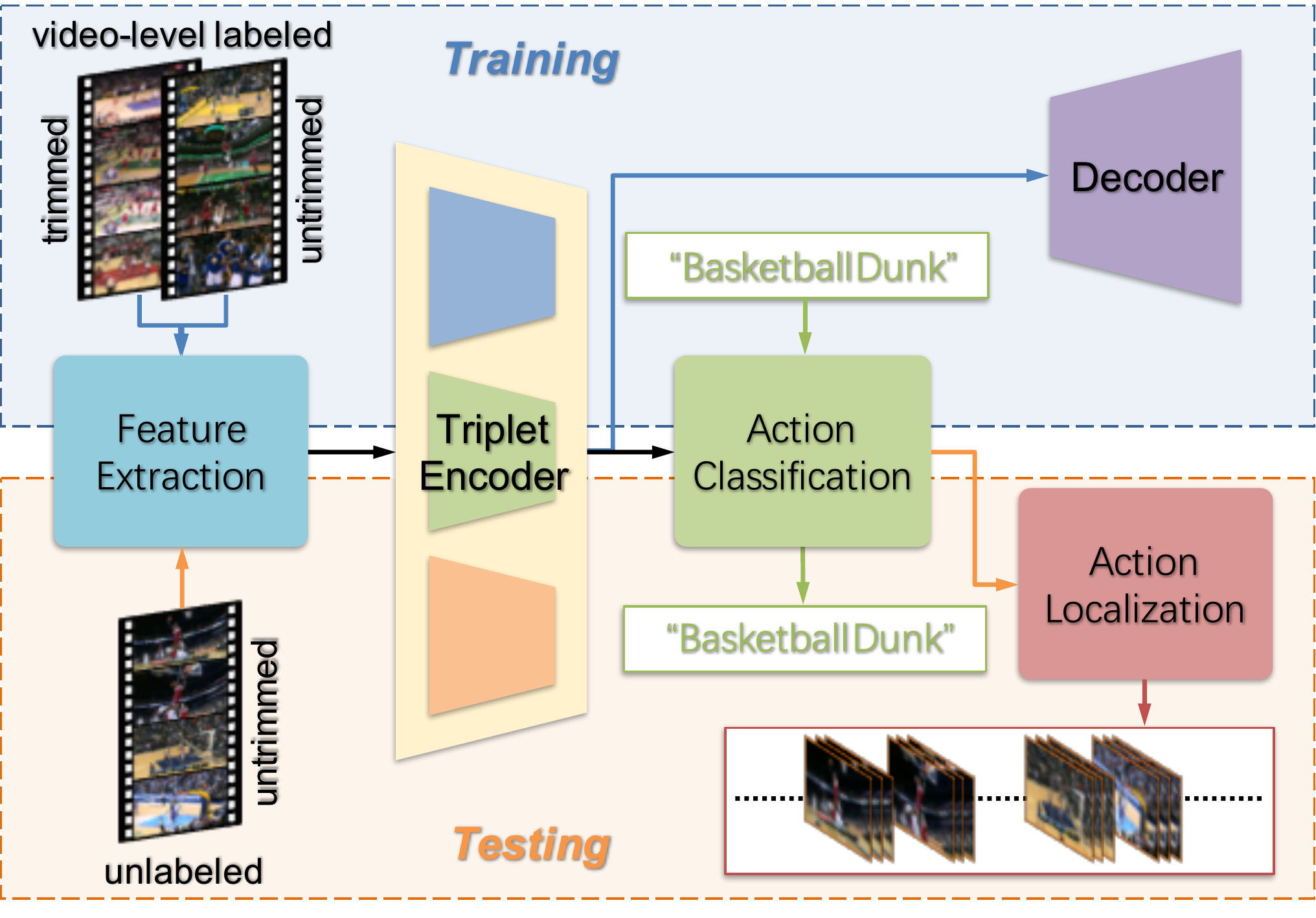}
\caption{The training-testing workflow of AdapNet. In the training stage, the method receives trimmed-untrimmed video pairs and the corresponding video-level labels, and learns the action recognition model through the encoder-decoder based network. In the testing stage, the optimized network is fed with unlabeled untrimmed videos, and aims to estimate both the actions of interest and their temporal intervals.}
\label{fig:1}
\end{figure}

To make up for the limited information available for untrimmed videos under weak supervision, a reasonable solution is to fully leverage trimmed videos as external resources to further improve the analytical performance. This is achieved by transferring the instructive knowledge learned on trimmed videos to untrimmed ones. In this way, the general latent patterns underlying trimmed and untrimmed videos of semantically related categories can be effectively shared via domain adaption. However, not all the aspects of video features are transferable. Overemphasizing the irrelevant aspects will inevitably jeopardize the generalization ability of the learning model~\cite{LJGR,EAOT,TaT,OnlineMultitask,RobustCost}. As a result, knowledge transfer without discrimination may bring about negative influence, and thus is not a reliable solution. To be specific, knowledge transfer should be carefully confined within the domain-adaptable features, whereas strictly exclude the domain-specific ones. 

As we know, it has become an emerging direction in machine learning to simultaneously exploit multiple aspects of the input feature space, which proves beneficial for various learning tasks. In this paper, we aim to decompose the original video features into domain-adaptable ones that are shared across different domains and domain-specific ones that are private to each domain, so as to enforce robust trimmed-untrimmed knowledge transfer. The shared-private decomposing mechanism is based on the sub-features’ adaptability during cross-domain modeling. Specifically, we propose a novel weakly supervised action recognition and localization method based on adaptability decomposing encoder-decoder deep neural network, referred to as AdapNet, whose training-testing workflow is illustrated in Figure~\ref{fig:1}. Throughout the procedure, both spatial and temporal relevance of frames in trimmed and untrimmed videos are explored effectively. 

The main contributions of the proposed method are summarized as follows.

\begin{itemize}
  \item We present adaptability decomposing of the original features to ensure reliable knowledge transfer between trimmed and untrimmed videos, so that domain adaptation can be safely confined within a more coherent subspace and instructive knowledge can be effectively shared.
  \item We develop an encoder-decoder based network structure to facilitate model learning, integrating a triplet encoder to extract domain-adaptable and domain-specific features and a decoder for information preservation.
  \item We design a bidirectional point process model by forward and backward exploring of video frames to achieve fine-grained temporal boundary detection, so that the time intervals corresponding to actions of interest can be identified precisely. 
  \item We conduct extensive experiments on two challenging untrimmed video datasets (i.e. THUMOS14~\cite{THUMOS14} and ActivityNet1.3~\cite{activitynet}), which show promising results of AdapNet over the existing state-of-the-art competitors.
\end{itemize}

The rest of this paper is organized as follows. We review the related work in Section II and introduce details of the proposed method in Section III. The results of experimental evaluation are reported in Section IV, followed by conclusions in Section V.

\section{Related Work}

\subsection{Action Recognition}
Video-based action recognition aims to identify the categories of actions in real-world videos. Many action recognition approaches have been proposed in the past few years. The improved dense trajectories (iDT)~\cite{idt1,idt2} has achieved the most outstanding performance as a hand-crafted feature. With the development of deep learning, tremendous progress has been made~\cite{MMC,EFC,HAR,UDL,Autoencoding}. For instance, two-stream network~\cite{two-stream1,two-stream2} is utilized to learn both appearance and motion information on frames and stacked optical flow respectively. C3D networks~\cite{c3d} adopt 3D convolutional kernel to capture both the spatial and temporal information directly from the raw videos. Wang et al.~\cite{tsn} designed a temporal segment network (TSN) to perform space sparse sampling and temporal fusion, which is aimed to learn from the integral videos. In addition, RNN methods are also widely used in action recognition tasks to improve the performance. Similar to the image classification tasks, which can be used in image object detection, action recognition models can also be used in temporal action localization tasks for feature extraction.
\subsection{Action Localization}
Different from action recognition, temporal action localization tries to identify the start and end time of actions of interest as well as recognizing the action categories for untrimmed videos. Previous works on temporal action localization mainly use the sliding windows as candidates and focus on designing hand-crafted feature representations to classify them~\cite{slidingwindow}. Recently, more and more works incorporate deep neural networks into the frameworks to obtain improved localization results. S-CNN~\cite{scnn} proposes to use multi-stage CNN to enhance the localization accuracy. Structured segment network (SSN)~\cite{ssn} proposes to model the temporal structure of each action instance via a structured temporal pyramid. SSAD~\cite{ssad} is proposed to use 1D temporal convolutional layers to skip the proposal generation step via directly detecting action instances in untrimmed videos. There are also some deep networks generating temporal proposals by probability estimation, such as Boundary Sensitive Network~\cite{BSN}. In addition, RNN is also widely used for temporal action localization. Compared with fully supervised methods, weakly supervised action localization is less studied. UntrimmedNets~\cite{untrimmednets} employs attention mechanisms to learn the pattern of precut action segments. STPN~\cite{STPN} utilizes a sparsity constraint to detect the activities, which improves the performance of action localization. TSRNet~\cite{TSRNet} integrates self-attention and transfer learning with temporal localization framework to obtain precise temporal intervals in untrimmed videos. AutoLoc~\cite{autoloc} is proposed to directly predict the temporal boundary of each action instance with an outer-inner-contrastive loss to train the boundary predictor. W-TALC~\cite{w-talc} learns the specific network weights by optimizing two complimentary loss functions, namely co-activity similarity loss and multiple instance learning loss. To fully leverage the publicly available trimmed videos, this paper further studies a reliable knowledge transfer mechanism from trimmed to untrimmed videos under adaptability constraint for effective action recognition and localization with weak supervision. 

\begin{figure*}[htb]
\centering
\includegraphics[width=1\linewidth]{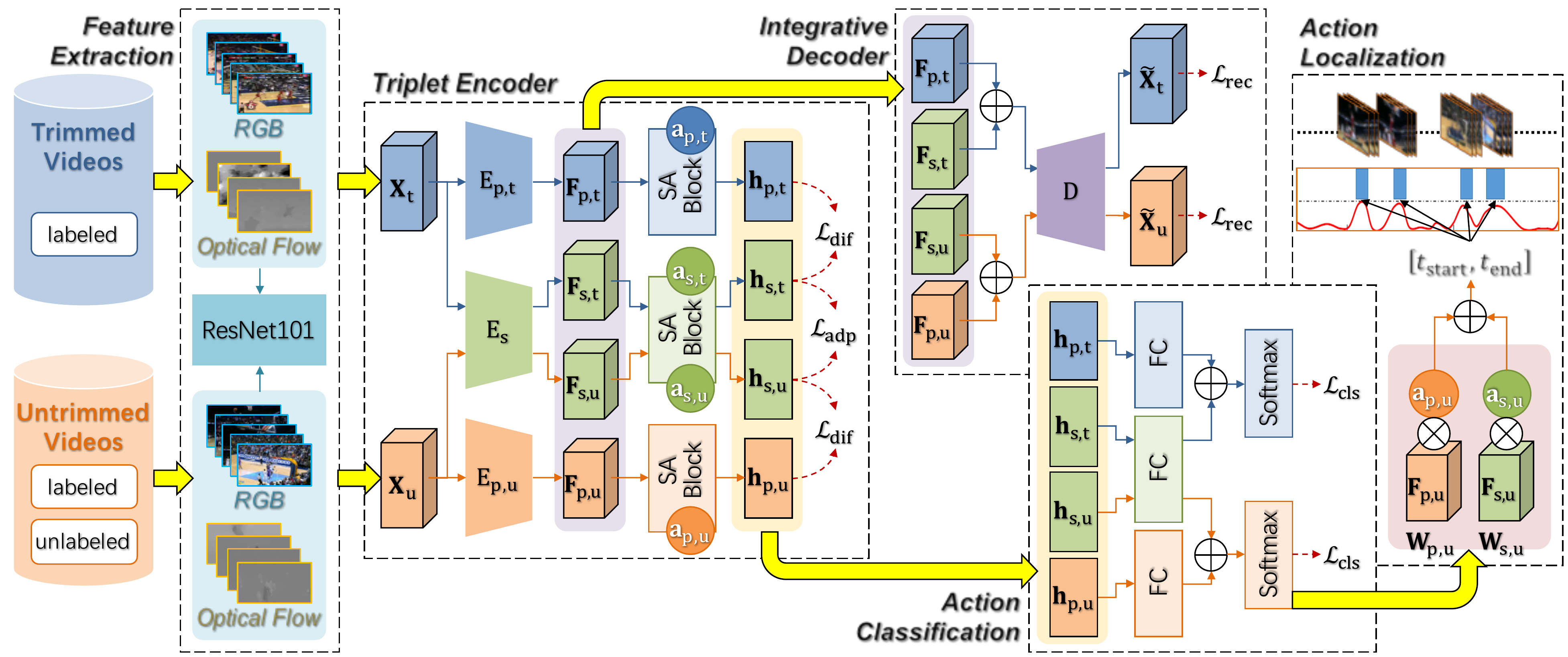}
\caption{The detailed framework of AdapNet (better viewed in color). The model starts with two\-stream feature extraction to fully represent the spatial and temporal properties of an input video. Then triplet encoder $\{\mathrm{E}_{\mathrm{p,t}}, \mathrm{E}_\mathrm{s}, \mathrm{E}_{\mathrm{p,u}}\}$ is implemented to decompose the features into domain-adaptable and domain-specific ones to ensure a reliable knowledge transfer from trimmed to untrimmed videos. An integrative decoder takes as input the decomposed features, and learns to recover the original video features with minimum information loss. Finally, the attention blocks corresponding to domain-adaptable and domain-specific features collaborate in an ensemble fashion for action recognition, based on which temporal localization can be subsequently deduced.}
\label{fig:2}
\end{figure*}

\subsection{Transfer Learning}
Transfer learning is aimed to learn a pattern that generalizes across different domains which are following different probability distributions. It is widely used in computer vision and natural language processing tasks. The main technical problem of transfer learning is how to reduce the domain shifts in data distributions. The most widely-used metric criterion on measuring the distance between different distributions is MMD (maximum mean discrepancy)~\cite{mmd}. Most existing methods are devoted to learning a shallow representation model through minimizing the domain discrepancy. However, the learned shallow representation model cannot suppress domain-specific exploratory factors of variations. With the rapid development of deep neural networks, the abstract representation learned by deep networks that disentangle the explanatory factors of variations behind data~\cite{DBLP:journals/pami/BengioCV13} and extract transferable factors underlying different populations~\cite{DBLP:conf/icml/GlorotBB11,DBLP:conf/cvpr/OquabBLS14} can only reduce the cross-domain discrepancy~\cite{DBLP:conf/nips/YosinskiCBL14}. There are also some works on domain adaptation choose to integrate the domain adaptation modules with deep neural networks to boost the transfer performance~\cite{DBLP:journals/corr/TzengHZSD14,DBLP:conf/iccv/TzengHDS15,DBLP:conf/icml/GaninL15,DBLP:conf/icml/LongC0J15,DBLP:conf/nips/LongZ0J16}, which can correct the marginal distributions shifts. 

\subsection{Feature Space Factorization}
Many computer vision problems inherently involve multiple views, where a view is broadly defined as a source or a representation of the data. Exploiting multiple views of information has proven beneficial for various practical applications. Recent studies indicate that multi-view learning is particularly effective when the views are either independent or fully dependent. Therefore, many methods attempt to factorize the original information and learn separate subspaces for modeling shared and private aspects of the data, which are designed with either linear mappings~\cite{ArchambeauB08,KlamiK08} or non-linear mappings~\cite{ek}. Salzmann et al.~\cite{SalzmannEUD10} proposed to encourage the shared-private factorization to be non-redundant while simultaneously discovering the dimensionality of the latent space. Jia et al.~\cite{jia} propose an approach to learning such factorized representations inspired by sparse coding techniques, which allows latent dimensions shared between any subset of the views instead of between all the views only. This paper is inspired by the shared-private factorization mechanism, and attempts to develop an effective video feature decomposing approach. 

\subsection{Encoder-decoder}
Encoder-decoder~\cite{cho2014learning} model is firstly proposed to solve the problem of seq2seq~\cite{DBLP:conf/nips/SutskeverVL14} to learn and describe latent attributes of the input data. The encoder is utilized to convert the input sequences to a vector whose length is fixed. The decoder is to convert the fixed vector to a sequence. There appears various variations of encoders, such as autoencoders~\cite{DBLP:journals/jmlr/Baldi12}. Autoencoders are an unsupervised learning tachnique in which the neural networks are leveraged for the task of representation learning. Specifically, autoencoder-decoder can convert the original input to a compressed knowledge representation, and measure the differences between the original input and the consequent reconstruction.
\section{Proposed Method}
In this section, we present the proposed AdapNet in detail, which consists of five modules, i.e. feature extraction, encoder, decoder, action classification and temporal localization modules. Figure~\ref{fig:2} illustrated the methodological framework of AdapNet.
\subsection{Spatio-temporal feature extraction}
As we know, videos naturally contain spatial and temporal components. In order to fully capture both spatial and temporal properties of a video, we adopt the widely used two-stream architecture for feature extraction. As the name suggests, video features are extracted via two separately trained networks corresponding to RGB and optical flow, respectively. We utilize ResNet101 pretrained on ImageNet for both streams, and receive frame-level video features. Formally, let $\mathcal{V}_*=\{f_i|_{i=1}^m\}$ stand for a video consisting of $m$ frames, where $f_i$is the $i$-th frame and $* \in \{\mathrm{t,u}\}$ indicates whether the video is trimmed or untrimmed. The RGB and optical flow features for $\mathcal{V}_*$ are denoted as $d$-by-$m$ matrices $\bm{\mathrm{X}}_{*,\mathrm{RGB}}$ and $\bm{\mathrm{X}}_{*,\mathrm{FLOW}}$ respectively, where $d$ is the dimension of feature vector for each frame $f_i \in \mathcal{V}_*$. Since both streams undergo identical subsequent procedures, without loss of generality, we use $\bm{\mathrm{X}}_*$ to denote the feature of $\mathcal{V}_*$ from a single stream. 

\begin{figure}[t]
\centering
\subfigure[]{\includegraphics[width=1\linewidth]{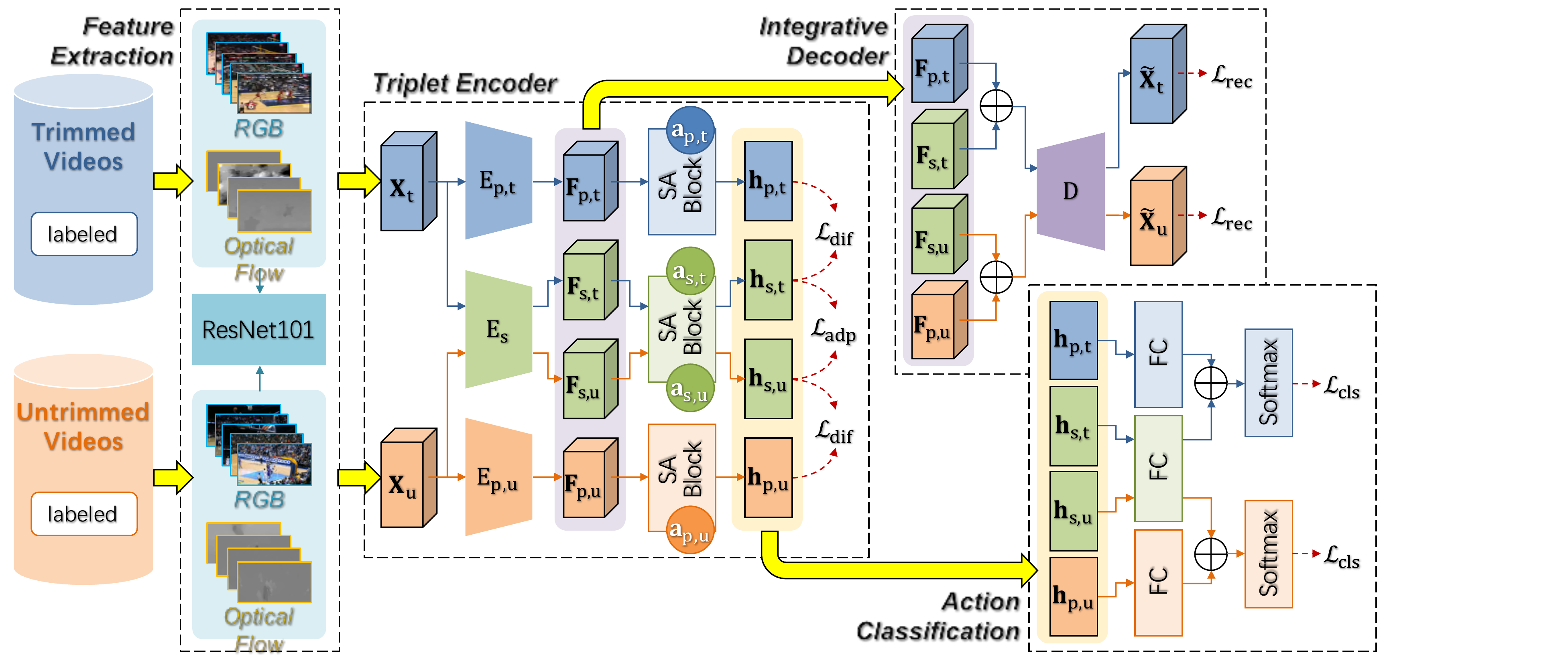}}
\subfigure[]{\includegraphics[width=1\linewidth]{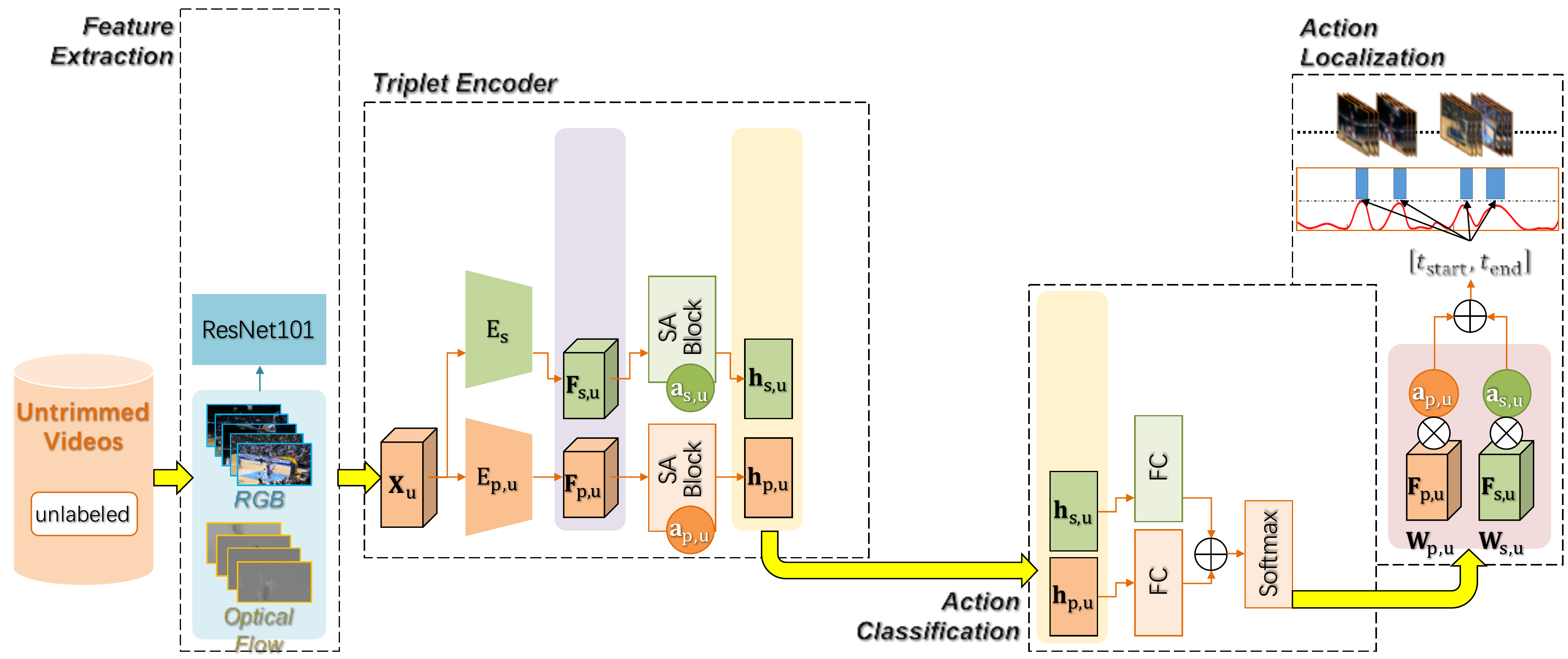}}
\caption{The activation of AdapNet in the (a) training and (b) testing stage.}
\label{fig4}
\end{figure}

\subsection{Triplet encoder for adaptability decomposing}
As a simplified version of real-life video, each trimmed video is well-aligned with the specific action of interest, and thus is ideal to develop effective action classification models. Apart from that, since the trimmed videos are not only semantically annotated but also temporally localized, they can also be invaluable resources for action recognition and localization of the untrimmed videos. Regarding the trimmed and untrimmed videos as source and target domains respectively, AdapNet aims to transfer the discriminative ability learned on trimmed videos to that on untrimmed ones. To achieve this, we propose adaptability decomposing of the features into domain-adaptable and domain-specific ones, so that domain adaption can be safely confined within a more coherent feature subspace.

We introduce the triplet encoder $\{\mathrm{E}_{\mathrm{p,t}}, \mathrm{E}_\mathrm{s}, \mathrm{E}_{\mathrm{p,u}}\}$, which consists of a shared encoder and two private encoders. The shared encoder $\mathrm{E}_\mathrm{s}$ focuses on the general features among semantically related trimmed-untrimmed video pairs, whereas the private encoder $\mathrm{E}_\mathrm{p,*} (* \in (\mathrm{t,u}))$ captures the exclusive features that dominate trimmed or untrimmed videos. Given a trimmed-untrimmed video pair $(\bm{\mathrm{X}}_{\mathrm{t}},\bm{\mathrm{X}}_{\mathrm{u}})$, the notations for decomposed features generated by triplet encoder are $(\bm{\mathrm{F}}_{\mathrm{s,t}},\bm{\mathrm{F}}_{\mathrm{s,u}})$ for the domain-adaptable ones and $(\bm{\mathrm{F}}_{\mathrm{p,t}},\bm{\mathrm{F}}_{\mathrm{p,u}})$ for the domain-specific ones. The former should be consistent with the semantic correlation of the videos to guarantee reliable knowledge transfer, and the latter should minimize the relevance to their counterparts. Note that both the decomposed features of a video $\mathcal{V}_*$ are matrices of the same size, i.e. $\bm{\mathrm{F}}_{\circ,*} \in \mathbb{R}^{d_{\mathrm{f}} \times m}$, where $\circ \in \{\mathrm{s,p}\}$ indicates the domain-adaptable or domain-specific features, and $d_{\mathrm{f}}$ is the dimension of decomposed feature vector for each frame.

\begin{table}[t]
\centering
\caption{Action recognition accuracy (\%) of state-of-the-art methods and AdapNet on THUMOS14.}
\begin{tabular}{p{5.8cm}|p{1.2cm}<{\centering}}
\hline
Method&Accuracy\\
\hline
Wang and Schmid, 2013~\cite{idt2}& 63.1 \\
Simonyan and Zisserman, 2014~\cite{two-stream2}& 66.1 \\
Jain et al., 2015~\cite{objectsmotion}&71.6 \\
Varol et al., 2015~\cite{extreme} & 63.2 \\
Zhang et al., 2016~\cite{emv}& 61.5 \\
Wang et al., 2016~\cite{tsn}&67.7 \\
Wang et al., 2016~\cite{tsn}&78.5 \\
Wang et al., 2017~\cite{untrimmednets}& 74.2 \\
Wang et al., 2017~\cite{untrimmednets}& 82.2 \\
Paul et al., 2018~\cite{w-talc}&85.6 \\
Zhang et al., 2019~\cite{TSRNet}&87.1\\
AdapNet& \textbf{87.9}\\
\hline
\end{tabular}
\label{table:table1}
\end{table}

\begin{table}[tp]
\centering
\caption{Action recognition accuracy (\%) of AdapNet with different implementations on ActivityNet1.3.}
\begin{tabular}{p{3.5cm}|p{0.8cm}<{\centering}p{1cm}<{\centering}p{1.8cm}<{\centering}}
\hline
& &Accuracy\\
\cline{2-4}
\multirow{1}*{Method}&\multirow{2}*{RGB}&Optical Flow&\multirow{2}*{Two-Stream}\\
\hline
Zhang et al., 2019 ~\cite{TSRNet}&79.7&84.3&91.2\\
\hline
AdapNet w/o $\mathcal{L}_{\mathrm{adp}}, \mathcal{L}_{\mathrm{dif}}$&71.9&74.8&80.4\\
AdapNet w/o $\mathcal{L}_{\mathrm{adp}}$&77.2&82.5&87.7\\
AdapNet w/o $\mathcal{L}_{\mathrm{dif}}$&81.6&85.9&90.2\\
AdapNet&\bf{83.4}&\bf{86.2}& \textbf{92.0}\\
\hline
\end{tabular}
\label{table:table2}
\end{table}

The decomposed features with variable length are then fed into Self-Attention (SA) blocks to further obtain fixed size embeddings. By learning a linear combination of the $m$ feature vectors of the frames, the SA block outputs the corresponding vector representations $\bm{\mathrm{h}}_{\mathrm{\circ,*}} \in \mathbb{R}^{d_{\mathrm{f}} \times 1}$. Formally, $\bm{\mathrm{h}}_{\mathrm{\circ,*}}$ is calculated as:
\begin{equation}
\bm{\mathrm{h}}_{\mathrm{\circ,*}} = \bm{\mathrm{F}}_{\mathrm{\circ,*}}\bm{\mathrm{a}}_{\mathrm{\circ,*}}.
\end{equation}
where
\begin{equation}\label{equ:2}
\bm{\mathrm{a}}_{\mathrm{\circ,*}} = \left(\mathrm{softmax}\left(\bm{\mathrm{w}}_2 \mathrm{tanh}\left(\bm{\mathrm{W}}_1\bm{\mathrm{F}}_{\mathrm{\circ,*}}\right)\right)\right)^\top.
\end{equation}
is a $m$-dimensional vector of attention weights. In~(\ref{equ:2}), $\bm{\mathrm{W}}_1 \in \mathbb{R}^{b \times d_{\mathrm{f}}}$ and $\bm{\mathrm{w}}_2 \in \mathbb{R}^{1 \times b}$ are intermediate parameters to be learned, where $b$ is a hyperparameter set empirically.

Let $\mathcal{T}=\{\mathcal{V}_{\mathrm{t}}^{(i)}|_{i=1}^{n_{\mathrm{t}}}\}$ and $\mathcal{U}=\{\mathcal{V}_{\mathrm{u}}^{(i)}|_{i=1}^{n_{\mathrm{u}}}\}$ be the sets of trimmed and untrimmed sets containing $n_{\mathrm{t}}$ and $n_{\mathrm{u}}$ videos, respectively. As a result, there are altogether $n_{\mathrm{t}} \times n_{\mathrm{u}}$ trimmed-untrimmed video pairs with video-level labels. By stacking the $d_{\mathrm{f}}$-dimensional video feature vectors, we arrive at the matrix format for features of the trimmed and untrimmed sets as $\bm{\mathrm{H}}_{\mathrm{\circ,t}}=[\bm{\mathrm{h}}_{\mathrm{\circ,t}}^{i}|_{i=1}^{n_\mathrm{t}}] \in \mathbb{R}^{d_{\mathrm{f}} \times n_{\mathrm{t}}}$ and $\bm{\mathrm{H}}_{\mathrm{\circ,u}}=[\bm{\mathrm{h}}_{\mathrm{\circ,u}}^{i}|_{i=1}^{n_\mathrm{u}}] \in \mathbb{R}^{d_{\mathrm{f}} \times n_{\mathrm{u}}}$, respectively. The loss functions of shared and private encoders are defined with Jensen-Shannon (JS) divergence as follows.
\begin{equation}
\mathcal{L}_{\mathrm{adp}}= \frac{1}{2}\mathrm{JS}(\bm{\mathrm{H}}_{\mathrm{s,t}}^{(\Omega)}||\bm{\mathrm{H}}_{\mathrm{s,u}}^{(\Omega)})-\frac{1}{2}\mathrm{JS}(\bm{\mathrm{H}}_{\mathrm{s,t}}^{(\bar{\Omega})}||\bm{\mathrm{H}}_{\mathrm{s,u}}^{(\bar{\Omega})}).
\end{equation}
\begin{equation}
\mathcal{L}_{\mathrm{dif}}= -\frac{1}{2}\mathrm{JS}(\bm{\mathrm{H}}_{\mathrm{p,t}}||\bm{\mathrm{H}}_{\mathrm{s,t}})-\frac{1}{2}\mathrm{JS}(\bm{\mathrm{H}}_{\mathrm{p,u}}||\bm{\mathrm{H}}_{\mathrm{s,u}}).
\end{equation}
where
\begin{equation}\label{equ:5}
\Omega = \{(i,j)|y_{\mathrm{t}}^{(i)}=y_{\mathrm{u}}^{(j)}\}.
\end{equation}
is the index set for semantically correlated video pairs, and $\bar{\Omega}$ is the complementary set of $\Omega$. In~(\ref{equ:5}), $y_*^{(i)}$ stands for the class label of video $\mathcal{V}_*^{(i)}$. On one hand, $\mathcal{L}_{\mathrm{adp}}$ encourages similar domain-adaptable feature distribution if the trimmed and untrimmed videos fall into identical action categories, and different otherwise. On the other hand, $\mathcal{L}_{\mathrm{dif}}$ punishes redundancy between the private and shared encoders.

As for the network structure, the shared and private encoders are all designed with three convolutional layers, followed by a max pooling layer of each convolutional layer, and finally a fully connected layer.

\subsection{Integrative decoder for information preservation}
After adaptability decomposing, the domain-adaptable and domain-specific features, i.e. $\bm{\mathrm{F}}_{\mathrm{s,*}}$ and $\bm{\mathrm{F}}_{\mathrm{p,*}}$, are fed into an integrative decoder to recover the original video representations. As discussed in the previous subsection, the decomposed features focus on different aspects of the video, and thus can provide complimentary information to each other when integrated. The recovery performance reflects the information loss during feature decomposing. For the sake of information preservation, we define the recovery loss as follows.
\begin{equation}
\mathcal{L}_{\mathrm{rec}} = \frac{1}{2n_{\mathrm{t}}}\sum_{i=1}^{n_{\mathrm{t}}}||\bm{\mathrm{X}}_{\mathrm{t}}^{(i)}-\bm{\widetilde{\mathrm{X}}}_{\mathrm{t}}^{(i)}||_{\mathrm{F}}^2 +\frac{1}{2n_{\mathrm{u}}}\sum_{i=1}^{n_{\mathrm{u}}}||\bm{\mathrm{X}}_{\mathrm{u}}^{(i)}-\bm{\widetilde{\mathrm{X}}}_{\mathrm{u}}^{(i)}||_{\mathrm{F}}^2.
\end{equation}
where $\bm{\mathrm{X}}_*^{(i)}$ and $\bm{\widetilde{\mathrm{X}}}_*^{(i)}$ denote the original and recovered feature matrix of video $\mathcal{V}_*^{(i)}$, respectively. $||\cdot||_{\mathrm{F}}^2$ is Frobenius norm widely used in matrix recovery tasks. 

The decoder is designed with a fully connected layer, followed by two convolutional layers, one upsampling layer and two convolutional layers.

\subsection{Joint action classification}
To identify the actions in each video, we pass the self-attentive representations $\bm{\mathrm{h}}_{\mathrm{\circ,t}}$ into the action classification module, which consists of two parts, i.e. a shared and two private FC layers, followed by two softmax activation layers to generate probabilistic predictions on the trimmed and untrimmed videos respectively.

In the training stage, the trimmed-untrimmed video pairs with video-level action labels are leveraged to learn robust classification models by minimizing the classification loss $\mathcal{L}_{\mathrm{cls}}$, which is computed with the standard multi-label cross-entropy loss on both trimmed and untrimmed videos. Note that the shared FC layer is trained with domain-adaptable features derived from both video sources. In this way, the classifier’s discriminative ability can be transferred from the trimmed to the untrimmed videos. The outputs from shared and private FC layers are integrated before feeding into the softmax layer, and predictions can be made based on the probabilistic scores.

Given a training dataset, the three modules (i.e. encoder, decoder, and classification) are optimized jointly with the overall loss function defined as follows.
\begin{equation}
\mathcal{L} = \mathcal{L}_{\mathrm{cls}} + \alpha\mathcal{L}_{\mathrm{adp}} + \beta\mathcal{L}_{\mathrm{dif}} + \gamma\mathcal{L}_{\mathrm{rec}}.
\label{overallloss}
\end{equation}
where $\alpha$, $\beta$ and $\gamma$ are hyper-parameters that control the trade-off between different terms.

\begin{algorithm}[t]
\caption{AdapNet - Training}
\begin{algorithmic}[1]
\REQUIRE $\mathcal{T}=\{\mathcal{V}_{\mathrm{t}}^{(i)}|_{i=1}^{n_{\mathrm{t}}}\}$, $\mathcal{U}=\{\mathcal{V}_{\mathrm{u}}^{(i)}|_{i=1}^{n_{\mathrm{u}}}\}$: the trimmed and untrimmed video sets; $\mathcal{Y}_{\mathcal{T}}$, $\mathcal{Y}_{\mathcal{U}}$: the corresponding video-level labels; $n_{\mathrm{b}}$: batch size; $\alpha, \beta, \gamma$: hyper-parameters.
\ENSURE $\bm{\mathrm{W}}$: parameters of AdapNet.
\STATE Randomly initialize $\bm{\mathrm{W}}$.
\WHILE {$\bm{\mathrm{W}}$ not converged}
\STATE Sample $\mathcal{S} \subset \mathcal{T} \times \mathcal{U} = \{(\mathcal{V}_{\mathrm{t}}^{(i)}|_{i=1}^{n_{\mathrm{t}}}, \mathcal{V}_{\mathrm{u}}^{(j)}|_{j=1}^{n_{\mathrm{u}}})\}$: a batch of trimmed-untrimmed video pairs, where \\$|\mathcal{S}|=n_{\mathrm{b}}\ll|\mathcal{T} \times \mathcal{U}| = n_{\mathrm{t}} \times n_{\mathrm{u}}$.\\
\STATE Update $\bm{\mathrm{W}}$ by minimizing $\mathcal{L}$ on $\mathcal{S}$:\\
$\bm{\mathrm{W}} = \argmin\limits_{\bm{\mathrm{W}}} \mathcal{L}(\mathcal{S})$.
\ENDWHILE
\label{algorithm1}
\end{algorithmic}
\end{algorithm}

\begin{algorithm}[t]
\caption{AdapNet - Testing}
\begin{algorithmic}[1]
\REQUIRE $\mathcal{Z} = \{\mathcal{V}_{\mathrm{z}}^{(i)}|_{i=1}^{n_{\mathrm{z}}}\}$: the unlabeled untrimmed video sets; $\bm{\mathrm{W}}$: parameters of AdapNet.
\ENSURE $\mathcal{Y}_{\mathcal{Z}}$: the predicted video-level labels; $\mathcal{J}_{\mathcal{Y}_{\mathcal{Z}}}$: the corresponding time intervals.
\STATE Predict multi-class classification probability:\\
$\mathcal{P}_{\mathcal{Z}} = \mathrm{AdapNet}(\mathcal{Z}|\bm{\mathrm{W}})$.
\STATE Identify actions:\\
$\mathcal{Y}_{\mathcal{Z}} = \mathrm{Thresholding}(\mathcal{P}_{\mathcal{Z}})$.
\STATE Localize actions based on multi-view T-CAM, NMS, etc.:\\
$\mathcal{J}_{\mathcal{Y}_{\mathcal{Z}}} = \mathrm{Localization}(\mathcal{Z}, \mathcal{P}_{\mathcal{Z}}, \bm{\mathrm{W}})$.
\label{algorithm2}
\end{algorithmic}
\end{algorithm}

\begin{figure*}[htb]
\centering
\includegraphics[width=1\linewidth]{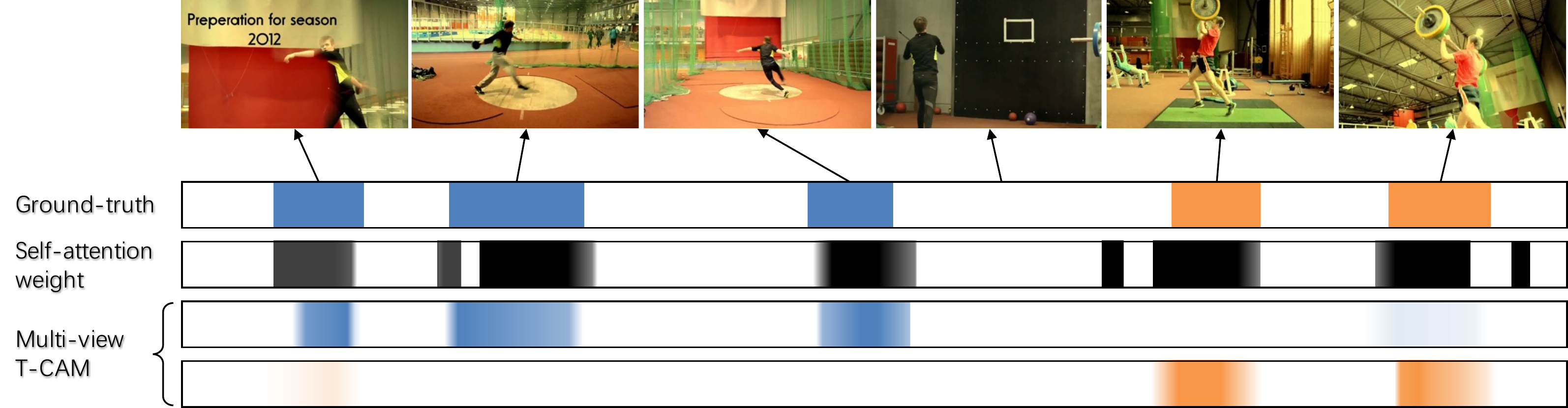}
\caption{Illustration of the temporal intervals corresponding to the ground-truth, the self-attention weights, and the multi-view T-CAM scores for a THUMOS14 video containing two action classes, i.e. $ThrowDiscus$ (Blue) and $CleanAndJerk$ (Orange). The horizontal axis is represented as the timestamps. As a class-agnostic metric, the self-attention weights highlight potential time intervals w.r.t. all the classes. In contrast, the multi-view T-CAM scores are calculated in a class-specific way, and thus provide accurate action localization information for $ThrowDiscus$ and $CleanAndJerk$, respectively.}
\label{fig:agnostic}
\end{figure*}

\subsection{Temporal action localization with bidirectional point process}
After model training, AdapNet can be used for both action classification and localization of untrimmed videos without labels. In this stage, the decoder module is no longer necessary, and only the networks related to untrimmed videos are involved. 

Following procedures similar to the training stage, actions can be predicted based on the probabilistic output of the softmax layer in the classification module. 

To further discern the temporal intervals in the untrimmed videos corresponding to the actions of interest, we fully explore the frame-level class-specific information. Inspired by the Temporal Class Activation Map (T-CAM), we propose the multi-view T-CAM to further fuse both domain-adaptable and domain-specific aspects of the video. 

Let $\bm{\mathrm{W}}_{\mathrm{\circ,u}} \in \mathbb{R}^{C \times d_{\mathrm{f}}}$ denote the weight parameter of the FC layer of the classification module, where $C$ is the number of classes. T-CAM, denoted by $\bm{\mathrm{P}}_{\mathrm{\circ,u}} \in \mathbb{R}^{C \times m}$, indicates the relevance of each frame to each class, whose element at $(c \in \{1,...,C\}, k \in \{1,...,m\})$ is defined as:
\begin{equation}
\bm{\mathrm{P}}_{\mathrm{\circ,u}}(c,k)= \bm{\mathrm{W}}_{\mathrm{\circ,u}}(c,:)\bm{\mathrm{F}}_{\mathrm{\circ,u}}(:,k).
\end{equation}
where $\bm{\mathrm{W}}_{\mathrm{\circ,u}}(c,:)$ is the $c$-th row of $\bm{\mathrm{W}}_{\mathrm{\circ,u}}$ and $\bm{\mathrm{F}}_{\mathrm{\circ,u}}(:,k)$ is the $k$-th column of $\bm{\mathrm{F}}_{\mathrm{\circ,u}}$ corresponding to the decomposed feature vector of the $k$-th frame. The attention weights can be further incorporated to obtain weighted T-CAM as:
\begin{equation}
\bm{\mathrm{Q}}_{\mathrm{\circ,u}}(c,k)=\bm{\mathrm{a}}_{\mathrm{\circ,u}}(k)\mathrm{sigmoid}(\bm{\mathrm{P}}_{\mathrm{\circ,u}}(c,k)).
\end{equation}
which takes into account both generic and class-specific information.

As we know, the domain-adaptable and domain-specific representations depict the video from different point of view. The multi-view T-CAM combines information from both view, which is defined as follows.
\begin{equation}
\bm{\mathrm{R}}_{\mathrm{u}}(c,k)=\delta\bm{\mathrm{Q}}_{\mathrm{s,u}}(c,k)+(1-\delta)\bm{\mathrm{Q}}_{\mathrm{p,u}}(c,k).
\end{equation}
where $\delta \in [0,1]$ is the parameter to control the significance of information from two views. Subsequently, multi-view T-CAM scores from the RGB and optical flow streams should also be fused so as to derive the two-stream multi-view T-CAM as: 
\begin{equation}
\bm{\mathrm{S}}_{\mathrm{u}}(c,k)=\varepsilon\bm{\mathrm{R}}_{\mathrm{u,RGB}}(c,k)+(1-\varepsilon)\bm{\mathrm{R}}_{\mathrm{u,FLOW}}(c,k).
\end{equation}
where $\varepsilon \in [0,1]$ is the balancing parameter.

Based on $\bm{\mathrm{S}}_{\mathrm{u}}$, smoothing is implemented to ensure the continuity of video contents. In order to obtain fine-grained temporal boundaries, we propose a bidirectional point process network for action probability estimation. Given two predefined thresholds, i.e. $\tau_{\mathrm{upper}}$ and $\tau_{\mathrm{lower}}$ ($\tau_{\mathrm{upper}} > \tau_{\mathrm{lower}}$), the frames whose $\bm{\mathrm{S}}_{\mathrm{u}}$ scores are above $\tau_{\mathrm{upper}}$ and below $\tau_{\mathrm{lower}}$ are attached with pseudo-labels as actions and backgrounds, respectively. By this manner, the bidirectional point process action estimator is trained in a semi-supervised fashion. As for the frames fall between $\tau_{\mathrm{upper}}$ and $\tau_{\mathrm{lower}}$, they are regarded as unlabeled frames to be estimated. To be specific, we leverage the twin recurrent point process (TRPP) model trained on the pseudo-labeled frames to estimate action probabilities of subsequent unlabeled frames. Furthermore, to model the preceding frames, we rearrange the video frames in reverse order, and learn another TRPP model. Finally, probabilities estimated with the bidirectional TRPP are fused into a comprehensive metric, i.e. the bidirectional point process refined multi-view T-CAM score.

After that, we perform Non-Maximum Suppression (NMS) to remove the highly overlapped predictions and arrive at the frame indices of starting and ending positions $[ind_{\mathrm{start}}, ind_{\mathrm{end}}]$, which should further be converted to the temporal intervals $[t_{\mathrm{start}}, t_{\mathrm{end}}]$ based on fps (frames per second).

\subsection{Summarization}
AdapNet takes on an encoder-decoder network structure that is a carefully designed for reliable trimmed-untrimmed knowledge transfer. The shared-private decomposing mechanism is enforced to guarantee coherent domain adaption.

To be specific, in the training stage, AdapNet is fed with trimmed-untrimmed video pairs and the corresponding video-level labels. According to the overall loss function in (\ref{overallloss}), the triplet encoder, integrative decoder, and action classification modules are jointly optimized in a unified procedure. In each training iteration, for the sake of implementation efficiency, we sample relatively subsets from the trimmed-untrimmed video pairs for network parameter update. 

After optimization, we deactivate the branches related to trimmed videos and the integrative decoder module. AdapNet in the testing stage receives unlabeled untrimmed videos as input to predict actions of interest. Action localization module is implemented to further identify the corresponding temporal intervals.

The activated parts of the AdapNet framework in the training and testing are illustrated in (a) and (b) of Fig.~\ref{fig4}, respectively. The training and testing procedures are summarized in Algorithm 1 and Algorithm 2.

\begin{table*}[htb]
\centering
\caption{Comparison of action localization results on THUMOS14.}
\begin{tabular}{c|l|ccccccccc}
\hline
\multirow{2}*{Supervision}&\multirow{2}*{Method}&\multicolumn{9}{c}{mAP@IoU (\%)}\\
\cline{3-11}
& &0.1&0.2&0.3&0.4&0.5&0.6&0.7&0.8&0.9\\
\hline
\multirow{7}*{Full}
&Richard and Gall, 2016~\cite{richard}&39.7&35.7&30.0&23.2&15.2&-&-&-&-\\
&Shou et al., 2016~\cite{scnn} &47.7&43.5&36.3&28.7&19.0&10.3&5.3&-&-\\
&Yeung et al., 2016~\cite{yeung}&48.9&44.0&36.0&26.4&17.1&-&-&-&-\\
&Yuan et al., 2016~\cite{yuan}&51.4&42.6&33.6&26.1&18.8&-&-&-&-\\
&Xu et al., 2017~\cite{r-c3d}&54.5&51.5&44.8&35.6&28.9&-&-&-&-\\
&Zhao et al., 2017~\cite{ssn}&\bf{66.0}&\bf{59.4}&51.9&41.0&29.8&-&-&-&-\\
&Chao et al., 2018~\cite{chao}&59.8&57.1&\bf{53.2}&\bf{48.5}&\bf{42.8}&\bf{33.8}&\bf{20.8}&-&-\\

\hline
\multirow{8}*{Weak}
&Singh and Lee, 2017~\cite{hideandseek}&36.4&27.8&19.5&12.7&6.8&-&-&-&-\\
&Wang et al., 2017~\cite{untrimmednets}&44.4&37.7&28.2&21.1&13.7&-&-&-&-\\
&Nguyen et al., 2018~\cite{STPN}&45.3&38.8&31.1&23.5&16.2&9.8&5.1&2.0&0.3\\
&Nguyen et al., 2018~\cite{STPN}&52.0&44.7&35.5&25.8&16.9&9.9&4.3&1.2&0.1\\
&Shou et al., 2018~\cite{autoloc}&-&-&35.8&29.0&21.2&13.4&5.8&-&-\\
&Paul et al., 2018~\cite{w-talc}&55.2&49.6&40.1&31.1&22.8&-&7.6&-&-\\
&Zhang et al., 2019\cite{TSRNet}&55.9&46.9&38.3&28.1&18.6&11.0&5.59&2.19&0.29\\
&AdapNet&\bf{56.51}&\bf{51.18}&\bf{41.09}&\bf{31.61}&\bf{23.65}&\bf{14.53}&\bf{7.75}&\bf{2.42}&\bf{0.36}\\
\hline
\end{tabular}
\label{thumos}
\end{table*}

\begin{table*}[htb]
\centering
\caption{Comparison of action localization results on the ActivityNet1.3.}
\begin{tabular}{c|l|cccc}
\hline
\multirow{2}*{Supervision}&\multirow{2}*{Method}&\multicolumn{4}{c}{mAP@IoU (\%)}\\
\cline{3-6}
& &0.5&0.75&0.95&Average\\
\hline
\multirow{5}*{Full}
&Xu et al., 2017~\cite{r-c3d}&26.8&-&-&-\\
&Heilbron et al., 2017~\cite{scc}&40.0&17.9&4.7&21.7\\
&Shou et al., 2017~\cite{cdc}&45.3&26.0&0.2&23.8\\
&Zhao et al., 2017~\cite{ssn}&39.12&23.48&5.49&23.98\\
&Lin et al., 2018~\cite{BSN}&\textbf{52.50}&\textbf{33.53}&\textbf{8.85}&\textbf{33.72}\\
\hline
\multirow{3}*{Weak}
&Nguyen et al., 2018~\cite{STPN}&29.3&16.9&2.6&-\\
&Zhang et al., 2019~\cite{TSRNet}&33.1&18.7&3.32&21.78\\
&AdapNet&\textbf{33.61}&\textbf{18.75}&\textbf{3.40}&\textbf{21.97}\\
\hline
\end{tabular}
\label{table:activitynet}
\end{table*}

\begin{table}[tp]
\centering
\caption{Comparison of action localization results of AdapNet with different implementations on THUMOS14 and ActivityNet1.3.}
\begin{tabular}{l|c|c}
\hline
\multirow{2}*{Method}&\multicolumn{2}{c}{mAP@IoU=0.5 (\%)}\\
\cline{2-3}
&THUMOS14&ActivityNet1.3\\
\hline
AdapNet w/o $\mathcal{L}_{\mathrm{adp}}, \mathcal{L}_{\mathrm{dif}}$ &6.26 & 17.24\\
AdapNet w/o $\mathcal{L}_{\mathrm{adp}}$ & 19.17 & 28.32\\
AdapNet w/o $\mathcal{L}_{\mathrm{dif}}$ & 20.82 & 30.11\\
AdapNet&\bf{23.65}&\bf{33.61}\\
\hline
\end{tabular}
\label{table:ablation}
\end{table}
\begin{figure*}[htbp]
\centering
\subfigure[An example of $ThrowDiscus$ (top) and $Shotput$ (down) actions.]{\includegraphics[width=1\linewidth]{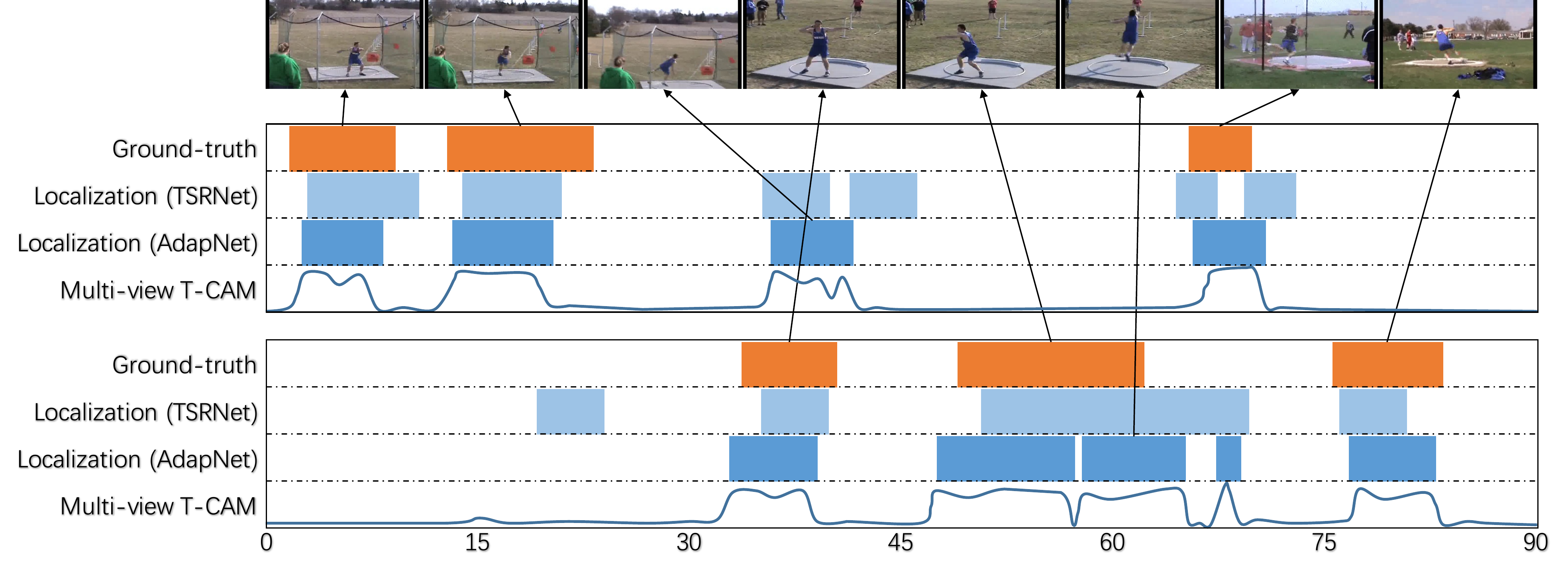}}
\subfigure[An example of $BasketballDunk$ action.]{\includegraphics[width=1\linewidth]{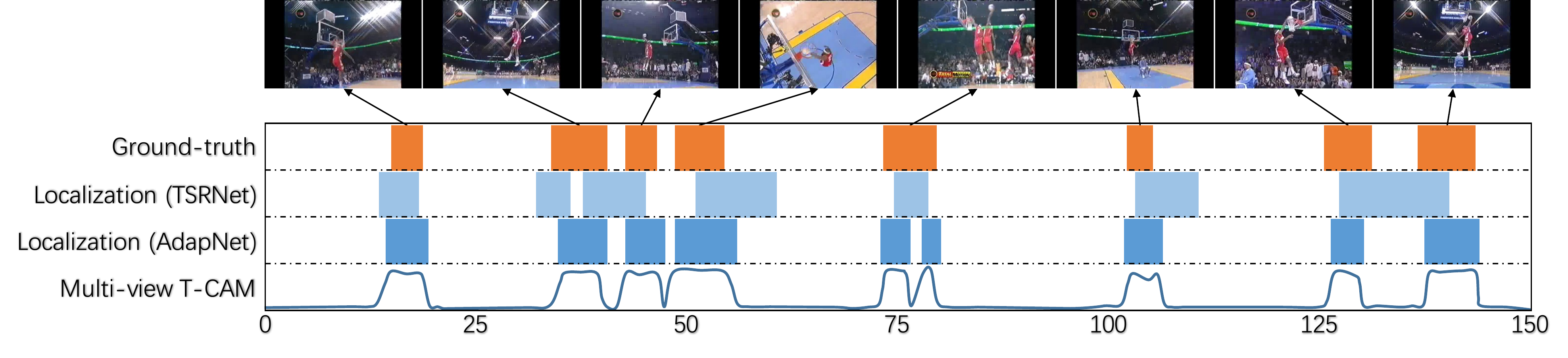}}
\subfigure[An example of $Surfing$ action.]{\includegraphics[width=1\linewidth]{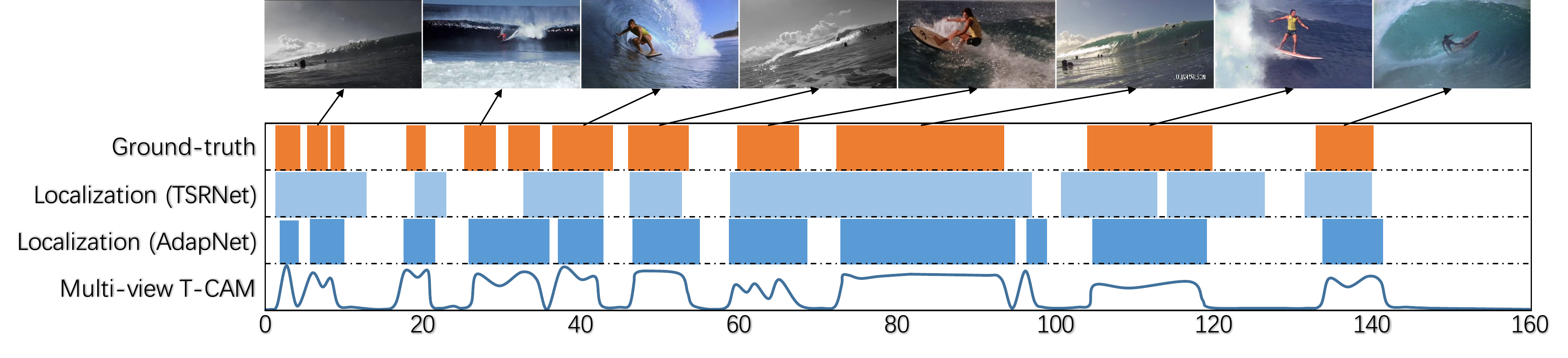}}
\caption{Qualitative results on THUMOS14, numbered (a) and (b), and ActivityNet1.3, numbered (c). The horizontal axis in the plots denote the timestamps (in seconds).}
\label{fig:loc}
\end{figure*}

\section{Experiments}
In this section, we report the experimental evaluation of the proposed AdapNet on benchmark datasets in comparison with other state-of-the-art algorithms based on both fully supervised and weakly supervised learning.

\subsection{Datasets}
Throughout the experiments, we evaluate the methods on two benchmark datasets, i.e. THUMOS14 and ActivityNet1.3. Both datasets contain large numbers of untrimmed videos, some of which are attached with temporal annotations of actions. Note that, as a weakly supervised method, AdapNet does not use the temporal annotations in the model training stage. 

The THUMOS14 dataset contains 101 action classes of video-level labeled videos, among which 20 classes are further attached with temporal annotations. Therefore, we only utilize the temporally labeled 20-class subset of THUMOS14. We train our model with the validation set which consists of 200 untrimmed videos, and evaluate the trained model with the testing set of 213 videos.

The ActivityNet1.3 dataset is originally comprised of 200 activity classes, with 10,024 videos for training, 4,926 for validation, and 5,044 for testing. Since the ground-truth labels for the original testing set are withheld, we adopt the training set for model training and the validation set for testing.

Besides untrimmed video sets, we also utilize a publicly available trimmed video set from UCF101 for knowledge transfer. The dataset contains 9,965 videos for training and 3,355 for testing, belonging to 101 action categories. We only use the training set for knowledge transfer.




\subsection{Implementation Details}
We initialize the network weights with pre-trained models from ImageNet dataset and to extract features for video frames with two-stream CNN networks. We use the mini-batch stochastic gradient descent algorithm to learn the network parameters, where the batch size is set to 32 and momentum set to 0.9. For spatial networks, we apply the TV-$L$1 algorithm to take 5-frame stacks optical flow as input and we set the learning rate to 0.001 and decreases every 3,000 iterations by a factor of 10. For temporal networks, we set the learning rate to 0.005 and decreases every 6,000 iterations by a factor of 10. For data augmentation, we use the techniques including horizontal flipping, corner cropping and so on. The parameters in the overall loss function, i.e. $\alpha$, $\beta$, and $\gamma$, are empirically evaluated as 0.5, 0.1, and 0.01, respectively. Our algorithm is implemented in PyTorch.

We follow the standard evaluation metric, which is based on the values of mean average precision (mAP) under different levels of intersection over union (IoU) thresholds.

\subsection{Action Recognition}
We first apply AdapNet to action recognition task on THUMOS14, in comparison with the state-of-the-art methods. The recognition accuracies are listed in Table~\ref{table:table1}. As we can see, methods leveraging auxiliary trimmed videos with trimmed-untrimmed knowledge transfer, i.e. TSRNet and AdapNet, achieve significant improvement over their counterparts. It indicates that the learning model can benefit a lot from the instructive knowledge derived from external resources. By further confining domain adaptation within a more coherent subspace, AdapNet ensures reliable knowledge transfer between trimmed and untrimmed videos and receives the highest recognition performance.

To the best of our knowledge, very few action recognition methods are conducted on ActivityNet1.3. As a result, we only list the available results of TSRNet in Table~\ref{table:table2}, and compare different versions of AdapNet accordingly. To be specific, AdapNet optimized without one or both of $\mathcal{L}_{\mathrm{adp}}$ and $\mathcal{L}_{\mathrm{dif}}$ are tested, to evaluate the contribution in the overall loss function. The accuracies of each single stream, i.e. RGB and optical flow, and the two-stream input are also compared. As shown in Table~\ref{table:table2}, AdapNet also outperforms TSRNet on ActivityNet1.3. The full implementation of AdapNet achieves the highest performance, indicating that both shared and private encoders are indispensable parts of the framework. By integrating both RGB and optical flow features, the accuracy can be effectively augmented.

\subsection{Action Localization}
As introduced above, AdapNet identifies time intervals of the actions of interest based on the frame-level T-CAM scores calculated in the action localization module. In order to validate the effectiveness of this action indicator, Fig.~\ref{fig:agnostic} intuitively illustrates an example of the self-attentive weights and the multi-view T-CAM scores of a video containing multiple action categories. It is observed that the time intervals corresponding to all the actions can be effectively highlighted as a whole by the self-attentive weights. However, they cannot distinguish different actions because of the class-agnostic property. In contrast, the multi-view T-CAM scores are more consistent with the ground-truths, because they are calculated in a class-specific way.

We carry out action localization tasks on THUMOS14 and ActivityNet1.3, and evaluate AdapNet in comparison with both fully supervised and weakly supervised methods. The experimental results are recorded in Table~\ref{thumos} and Table~\ref{table:activitynet}, in which methods in the upper and lower parts are with full and weak supervision, respectively. It is observed that AdapNet significantly surpasses its weakly supervised counterparts. It is especially encouraging to see that AdapNet even achieves comparable or better results with some fully supervised methods. 

We further compare different implementations of AdapNet to validate the efficacy of adaptability decomposing with triplet encoder structure. Similar to Table~\ref{table:table2}, AdapNet optimized without one or both of $\mathcal{L}_{\mathrm{adp}}$ and $\mathcal{L}_{\mathrm{dif}}$ are evaluated, the localization results of mAP@IoU=0.5 are recorded in Table~\ref{table:ablation}. We observe that, both shared and private encoders are indispensable to learn a coherent subspace in formulating reliable knowledge transfer. AdapNet with integrated overall loss achieves the best localization performance. 

We also demonstrate the qualitative results on THUMOS14 and ActivityNet1.3 in Fig.~\ref{fig:loc}. As we can see, using multi-view TCAM as an effective indicator, the proposed method is capable of locating actions of interest in untrimmed videos. To be specific, Fig.~\ref{fig:loc}(a) presents an example of a video containing two actions, i.e. $ThrowDiscus$ and $Shotput$. Although the visual appearances and motion patterns are quite similar, AdapNet can successfully identify and localize most of the actions, with only a few false positives. Fig.~\ref{fig:loc} (b) and (c) show single-action videos of $BasketballDunk$ and $Surfing$, respectively. Despite of the challenges of large variations in scale, viewpoint, and illumination conditions, AdapNet still generate satisfactory localization results. As for the failure cases, we found that most false positives correspond to very short video clips which are closely related to actions of interest to human eyes, such as the preparation stage or ending stage of certain actions. We also visualize qualitative localization results of the most competitive TSRNet in Fig.~\ref{fig:loc}. As we can see, unconstrained knowledge transfer in TSRNet inevitably induce false positives in localization results. Some background contents or irrelevant actions are erroneously identified as the actions of interest. In contrast, AdapNet can ensure legitimate trimmed-untrimmed domain adaption within a semantically coherent subspace, and meanwhile prohibit irrelevant transfer. Therefore, AdapNet generates more satisfactory localization performance. 

\section{Conclusion}
In this paper, we have proposed a robust knowledge transfer framework, namely AdapNet, for weakly supervised action recognition and localization in untrimmed videos via point process modeling. Through a carefully designed encoder-decoder based learning structure, adaptability decomposing is effectively implemented to guarantee legitimate trimmed-untrimmed domain adaption within a semantically coherent subspace, and meanwhile prohibit irrelevant transfer. Given the spatio-temporal descriptions of input videos, the triplet encoder decomposes the original features into domain-adaptable and domain-specific ones, whereas the integrative decoder is devised for feature recovery with minimal information loss. Accompanied with action classification and localization modules using compact self-attentive representations, the proposed method can not only predict actions but also identify the corresponding time intervals in untrimmed videos with only video-level labels. As demonstrated on two challenging untrimmed video datasets, AdapNet achieves superior performance over the state-of-the-art methods.
\ifCLASSOPTIONcaptionsoff
  \newpage
\fi



%

\bibliographystyle{IEEEtran}
\bibliography{tnnls}



%

\begin{IEEEbiography}[{\includegraphics[width=1in,height=1.25in,clip,keepaspectratio]{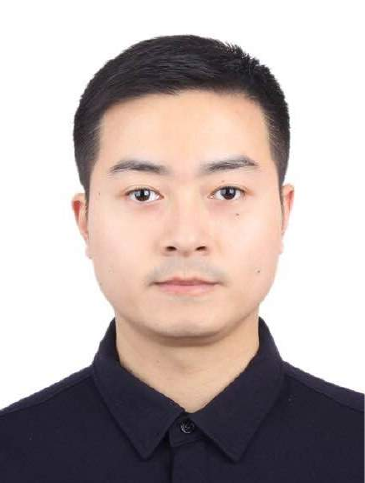}}]{Xiao-Yu Zhang}
received the B.S. degree in computer science from Nanjing University of Science and Technology, Nanjing, China, in 2005, and the Ph.D. degree in pattern recognition and intelligent systems from the Institute of Automation, Chinese Academy of Sciences, Beijing, China, in 2010. He is currently an Associate Professor with the Institute of Information Engineering, Chinese Academy of Sciences, Beijing, China. His research interests include artificial intelligence, data mining, computer vision, etc. Dr. Zhang’s awards and honors include the Silver Prize of Microsoft Cup IEEE China Student Paper Contest in 2009, the Second Prize of Wu Wen-Jun AI Science \& Technology Innovation Award in 2016, the CCCV Best Paper Nominate Award in 2017, and the Third Prize of BAST Beijing Excellent S\&T Paper Award in 2018.
\end{IEEEbiography}

\vspace{-10 mm}
\begin{IEEEbiography}[{\includegraphics[width=1in,height=1.25in,clip,keepaspectratio]{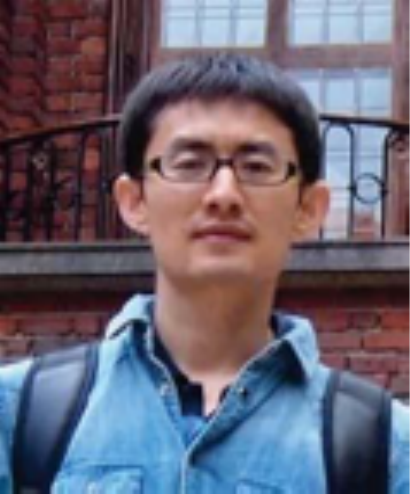}}]{Changsheng Li}
received the BE degree from the University of Electronic Science and Technology of China (UESTC), in 2008 and the PhD degree in pattern recognition and intelligent system from the Institute of Automation, Chinese Academy of Sciences, in 2013. He is currently a Full Research Professor from the University of Electronic Science and Technology of China. He is also a director of Youedata AI Research lab (YAIR). During pursuing his PhD, he once studied as a research assistant from The Hong Kong Polytechnic University in 2009-2010. After obtaining his PhD, he worked with IBM Research-China and iDST, Alibaba Group, respectively. His research interests include machine learning, data mining, and computer vision. He has more than 30 referred publications in international journals and conferences, including T-PAMI, T-IP, T-NNLS, T-C, PR, CVPR, AAAI, IJCAI, CIKM, MM, ICMR, etc.
\end{IEEEbiography}

\vspace{-10 mm}
\begin{IEEEbiography}[{\includegraphics[width=1in,height=1.25in,clip,keepaspectratio]{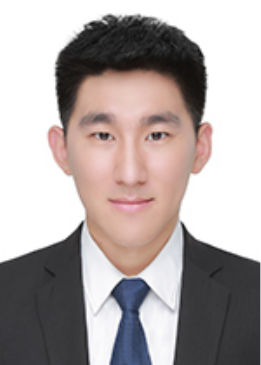}}]{Haichao Shi}
received the B.S. degree in software engineering from Beijing Technology and Business University, Beijing, China, in 2017. He is currently pursuing the M.S. degree for computer software and theory in National Engineering Laboratory for Information Content Security Technology, Institute of Information Engineering, Chinese Academy of Sciences, Beijing, China. His research interests include pattern recognition, image processing and video content analysis.
\end{IEEEbiography}

\vspace{-10 mm}
\begin{IEEEbiography}[{\includegraphics[width=1in,height=1.25in,clip,keepaspectratio]{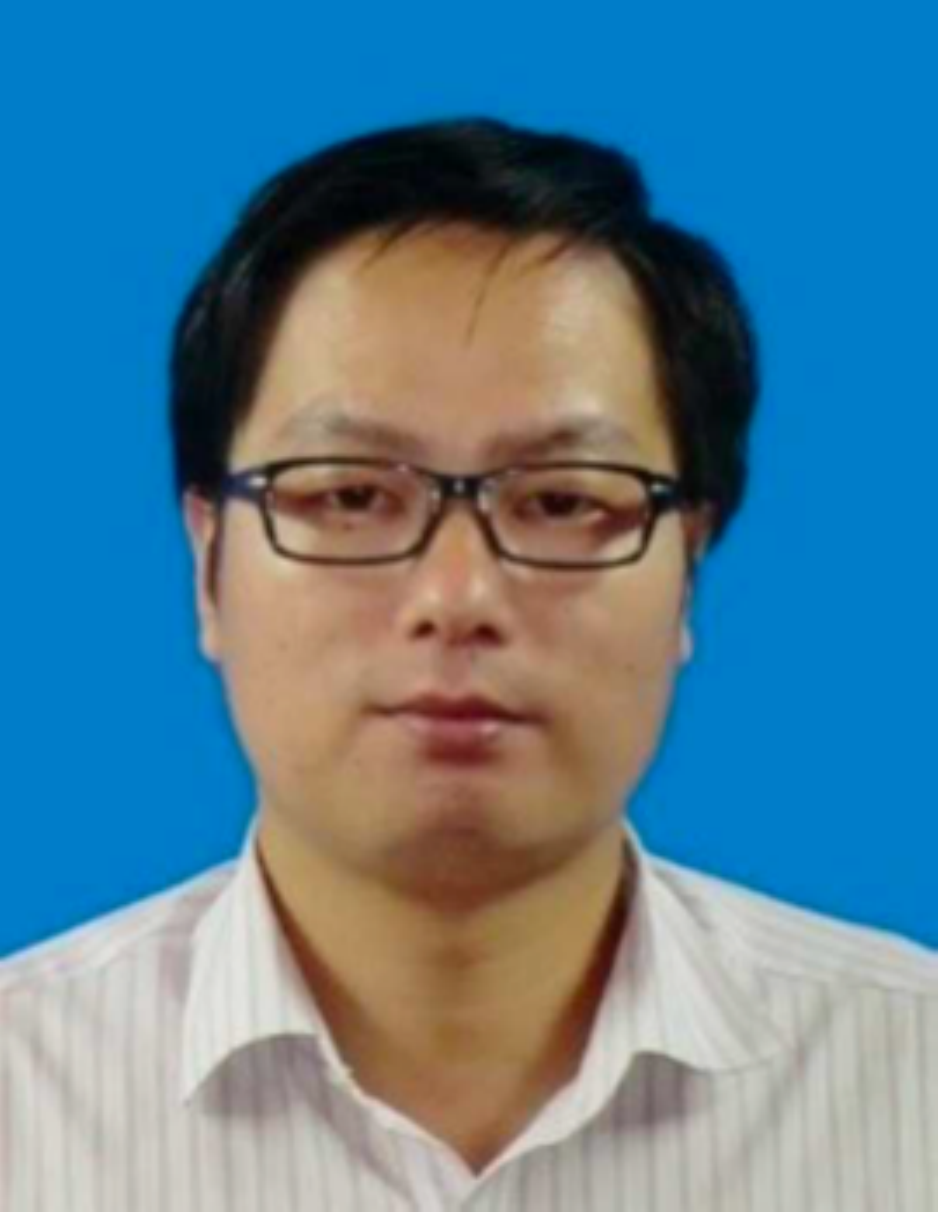}}]{Xiaobin Zhu}
received his M.E. degree in 2006 from Beijing Normal University, and his Ph.D. degree from Institute of Automation, Chinese Academy of Sciences in 2013. Currently he is an associate professor in School of Computer and Communication Engineering, University of Science and Technology Beijing. His research interests include machine learning, image content analysis and classification, multimedia information indexing and retrieval, etc.
\end{IEEEbiography}

\vspace{20 mm}
\begin{IEEEbiography}[{\includegraphics[width=1in,height=1.25in,clip,keepaspectratio]{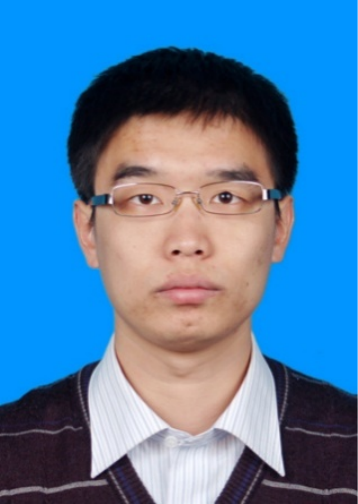}}]{Peng Li}
received the B.E. degree in automation from Shandong University, Jinan, China, in 2008, and the Ph.D. degree in pattern recognition and intelligent systems from the Institute of Automation, Chinese Academy of Sciences, Beijing, China, in 2013.
He is currently an Associate Professor with the College of Oceanography and Space Informatics, China University of Petroleum (East China), Qingdao, China. His research interests include machine learning methods and their applications in image processing and remote sensing. He was a recipient of Prize Paper Award Honorable Mention from IEEE Transactions on Multimedia in 2016.
\end{IEEEbiography}

\vspace{-130 mm}
\begin{IEEEbiography}[{\includegraphics[width=1in,height=1.25in,clip,keepaspectratio]{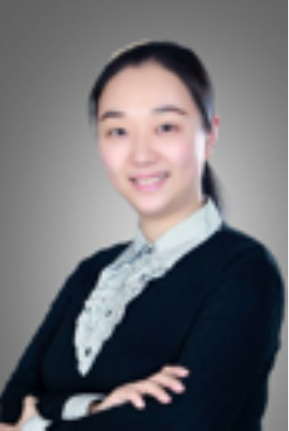}}]{Jing Dong}
received her BSc in Electronic Information Science and Technology from Central South University in 2005 and her PhD in Pattern Recognition from the Graduate University of Chinese Academy of Sciences. Since July 2010, she has joined the National Laboratory of Pattern Recognition (NLPR), where she is currently an Associate Professor. Her research interests include pattern recognition, image processing and digital image forensics. She has published over 30 academic papers and she is a member of CCF, CAAI, a Senior member of IEEE. She also serves as the deputy general of Chinese Association for Artificial Intelligence and the China Society of Image and Graphics. She also serves as an IEEE Volunteer leader in Region 10 and Beijing Section from many aspects of academic activities.
\end{IEEEbiography}




\end{document}